\documentclass[10pt,twocolumn,letterpaper]{article}

\usepackage{icb}
\usepackage{times}
\usepackage{epsfig}
\usepackage{graphicx}
\usepackage{amsmath}
\usepackage{amssymb}
\usepackage{color,soul}
\usepackage{balance}
\usepackage{siunitx}
\usepackage{multirow}

\usepackage{xcolor}
\usepackage[overlay]{textpos}

\icbfinalcopy 

\ificbfinal\fi

\begin{document}

\setlength{\abovedisplayskip}{3pt}
\setlength{\belowdisplayskip}{3pt}



\title{Semi-Adversarial Networks: Convolutional Autoencoders for Imparting Privacy to Face Images}


\author{Vahid Mirjalili\textsuperscript{ 1}  \qquad Sebastian Raschka\textsuperscript{ 1} \qquad Anoop Namboodiri~\textsuperscript{2}  \qquad Arun Ross\textsuperscript{ 1} \\ \;\;\;\;\;\; {\tt\small mirjalil@msu.edu} \;\;\;\; \; {\tt\small raschkas@msu.edu}  \;\;\;  \;\;\;   \;\; {\tt\small anoop@iiit.ac.in}  \;\;\;  {\tt\small rossarun@cse.msu.edu}  \\ \\
\textsuperscript{1}  Michigan State University, East Lansing, USA \\
\textsuperscript{2} International Institute of Information Technology, Hyderabad, India}



\maketitle
\thispagestyle{empty}

\begin{abstract}
In this paper, we design and evaluate a convolutional autoencoder that perturbs an input face image to impart privacy to a subject. Specifically, the proposed autoencoder transforms an input face image such that the transformed image can be successfully used for face recognition but not for gender classification. In order to train this autoencoder, we propose a novel training scheme, referred to as semi-adversarial training in this work. The training is facilitated by attaching a semi-adversarial module consisting of an auxiliary gender classifier and an auxiliary face matcher to the autoencoder.   The objective function utilized for training this network has three terms: one to ensure that the perturbed image is a realistic face image; another to ensure that the gender attributes of the face are confounded; and a third to ensure that biometric recognition performance due to the perturbed image is not impacted. Extensive experiments confirm the efficacy of the proposed architecture in extending gender privacy to face images. 
\end{abstract}

\begin{textblock*}{20cm}(0.0cm,-16cm) 
   \textcolor{red}{Published in \textit{Proc. of 11th IAPR International Conference on Biometrics (ICB 2018). Gold Coast, Australia, Feb. 2018}}
\end{textblock*}

\section{Introduction}

Biometric face recognition refers to the use of face images for recognizing an individual in an automated manner~\cite{jain_introduction_2011}. 
A typical face recognition system employs a {\em face matcher} that compares two face images and determines the degree of similarity or dissimilarity between them. This comparison operation can be used to (a) {\em verify} the claimed identity of an input face image  or (b) determine the {\em identity} of an unknown face  image by comparing it against a set of known face images. 

While face images collected  by a biometric system are expected to be used only for {\em recognition} of individuals~\cite{kindt_privacy_2016}, 
 recent research has established the possibility  of automatically deducing additional information about an individual from their face image~\cite{dantcheva_what_2016}. 
 For example, information about a person's age, gender, race, or health can be obtained by using a {\em soft biometric classifier} (\eg, a gender classifier) that can extract this information from a single face image ~\cite{sun_demographic_2017}. 
While the extraction of soft biometric data (sometimes referred to as {\em attributes}) can be used to improve the performance of a biometric system~\cite{rozsa_are_2016,madry_towards_2017}, it also raises several {\em privacy} concerns associated with gleaning information without an individual's consent. Further, such an automated analysis can be potentially misused for age-based or gender-based profiling that can undermine the use of biometrics in many applications~\cite{garvie_perpetual_2016}. 

Given these concerns, researchers have discussed the possibility of {\em de-identifying} a face image prior to storing it in a database~\cite{newton_preserving_2005}. 
While de-identification has tremendous applications in surveillance systems, it can irrevocably compromise the biometric utility of a face image~\cite{gross_integrating_2005}. 
However, in many applications, it is necessary to {\em retain} the biometric utility of the face image while {\em suppressing} the possibility of gleaning additional information, such as gender~\cite{mirjalili_soft_2017}. 
This type of {\em differential privacy}~\cite{othman_privacy_2014}
 is expected to enhance the privacy of face images stored in a database while at the same time  ensuring that biometric recognition is not unduly affected. 

In this work, we develop a convolutional autoencoder (CAE) that generates a perturbed face image that can be successfully used by a {\em face matcher} but not by a {\em gender classifier}. The proposed CAE is referred to as a {\bf semi-adversarial network} since its output is adversarial to the gender classifier but not to the face matcher. The proposed network can be easily appropriated for use with other attributes (such as age or race).  In principle, the design of the semi-adversarial network can be utilized in other problem domains where there is a need  to confound some classifiers while retaining the utility of other classifiers.

\subsection{Related work}


A number of aspects of privacy protection has been studied in the biometric literature~\cite{natgunanathan_protection_2016, ratha_enhancing_2001, othman_privacy_2014,mirjalili_soft_2017}. On one hand, there are face de-identification techniques~\cite{jourabloo_attribute_2015,newton_preserving_2005,gross_model-based_2006} where a face image is modified in order to confound a face matcher. On the other hand, as inspired by the work of Othman and Ross~\cite{othman_privacy_2014} and later promoted by Sim and Zhang~\cite{sim_controllable_2015}, the goal is to selectively confound or preserve a set of attributes that can be deduced from face images. Specifically, a few methods for suppressing the gender attribute have been presented~\cite{rowland_manipulating_1995,suo_high-resolution_2011,othman_privacy_2014}. Recently, a new method for protecting privacy with
practical applications for biometric databases was proposed
in~\cite{mirjalili_soft_2017}, where input face images were modified with respect
to a specific gender classifier. In this case, perturbations
were derived based on a specific gender classifier,
the perturbations did not significantly impact
the match scores of a face matcher.

In this paper, we provide an alternative solution by designing a convolutional autoencoder that transforms input images such that the performance of an {\em arbitrary} gender classifier is impacted, while that of an {\em arbitrary} face matcher is retained. The contributions of this paper, in this regard, are the following: (a) formulating the privacy-preserving problem in terms of a convolutional autoencoder that does {\em not} require prior knowledge about the gender classifier nor the face matcher being used; (b) incorporating an explicit term related to the matching accuracy in the objective function which ensures that the {\em utility} of the perturbed images is not negatively impacted;
(c) developing a {\em generalizable} solution that can be trained on one dataset and applied to other previously unseen datasets.

 
To the best of our knowledge, this is the first work where adversarial training is used to design a generator component that is able to maximize the performance with respect to one classifier while minimizing the performance with respect to another. Experimental results show that the proposed method of semi-adversarial learning for multi-objective functions is efficient for deriving perturbations that are generalizable to other classifiers that were not used (or not available) during training.
\section{Proposed method}

\subsection{Problem formulation} \label{sssec:num2}

Let $X \in \mathbb{R}^{m \times n \times c}$ denote a face image having $c$ channels each of height $m$ and width $n$. Let  $f_G(X)$ denote a binary gender classifier that  returns a value in the range $[0, 1]$, where $1$ indicates a ``Male'' and $0$ indicates a ``Female''. Let  $f_M(X_1, X_2)$ denote a face matcher that computes the match score between a pair of face images, $X_1$ and $X_2$. The goal of this work is to construct a model $\phi(X)$, that perturbs an input image $X$  such that the perturbed image $X'=\phi(X)$ has the following characteristics: (a) from a human perspective, the perturbed image $X'$ must look similar to the original input $X$;
(b) the perturbed image $X'$ is most likely to be misclassified by an arbitrary gender classifier $f_G(X)$;
(c) the match scores, as assessed by an arbitrary biometric matcher $f_M$, between perturbed image $X'$ and other unperturbed face images from the same subject, are not impacted thereby retaining verification accuracy.

This goal can be expressed as the following objective function, which minimizes a loss function $J$ consisting of three disjoint terms corresponding to the three characteristics listed above:
\begin{equation}
\begin{array}{cll}
J(X, y, X';  f_{G}, f_{M}) =& \\
  \lambda_D J_D(X, X') + \lambda_G J_G(y, X'; f_{G}) + 
 \lambda_M J_M(X, X'; f_{M}),
\end{array}
\label{eq:loss-function}
\end{equation}
where, $X$ is the input image, $y$ is the gender label of $X$, and $X'$ is the perturbed image. The term $J_D(X, X')$ measures the dissimilarity between the input image and the perturbed image produced by a decoder $\phi(X)$ to ensure that the perturbed images still appear as realistic face images. The second term, $J_G(y, X'; f_G)$, measures the loss associated with correctly predicting gender of perturbed image $X'$ using  $f_G$, to ensure that the accuracy of the gender classifier on the perturbed image $X^\prime$ is reduced. 
The third term, $J_M(X, X'; f_{M})$, measures the loss associated with the match score between $X$ and $X'$ computed by  $f_M$. This term ensures that the matching accuracy as assessed by $f_{M}$ is not  substantially diminished due to the perturbations introduced to confound the gender classifier.

In order to optimize this objective function, \ie, minimizing gender classifier accuracy while maximizing the biometric matching accuracy and generating realistic looking images, we design a novel convolutional neural network architecture that we refer to as a semi-adversarial convolutional autoencoder.

\subsection{Semi-adversarial network architecture}

\begin{figure}[h!]
\begin{center}
   \includegraphics[width=0.8\linewidth]{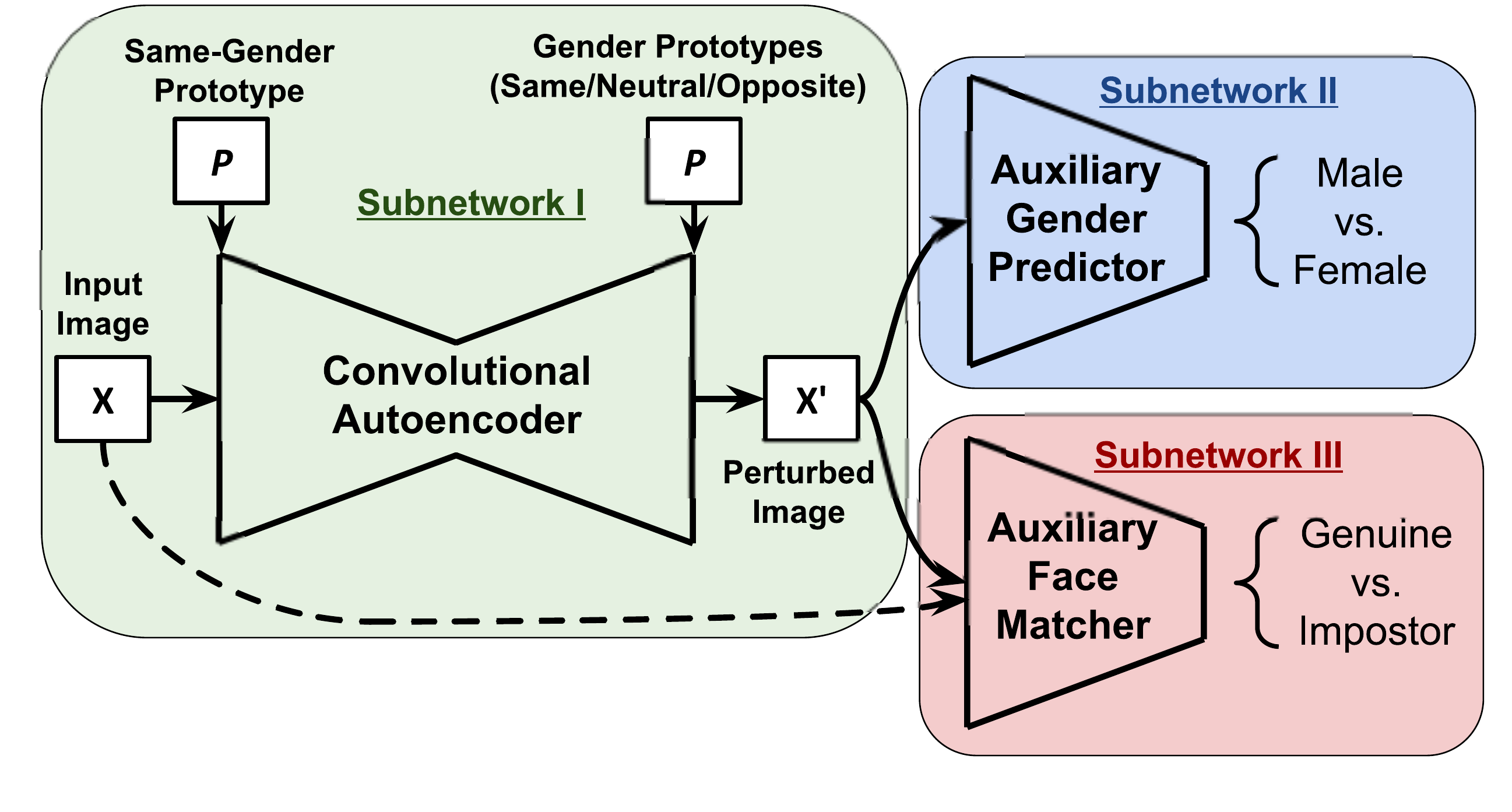}
\end{center}
   \caption{Schematic representation of the semi-adversarial neural network architecture designed to derive perturbations that are able to confound gender classifiers while still allowing biometric matchers to perform well. The overall network consists of three sub-components: a convolutional autoencoder (subnetwork I), an auxiliary gender classifier (subnetwork II), and an auxiliary matcher (subnetwork III).  }
\label{fig:general-arch}
\end{figure}

The semi-adversarial network introduced in this paper is significantly different from Generative Adversarial Networks (GANs). A typical GAN has two components: a discriminator and a generator.  The {\em generator} learns to generate realistic looking images from the training data, while the {\em discriminator} learns to distinguish between the generated images and the corresponding training data~\cite{goodfellow_generative_2014,rozsa_are_2016}. In contrast to regular GANs consisting of a generator and a single discriminator, the proposed semi-adversarial network attaches two independent classifiers to a generative subnetwork. Unlike the generator subnetwork of GANs that is trained based on the feedback of one classifier, the semi-adversarial configuration proposed in this paper learns to generate image perturbations based on the feedback of two classifiers, where one classifier acts as an adversary of the other. Hence, the semi-adversarial network architecture we propose consists of the following three different subnetworks (Fig.~\ref{fig:general-arch}): (a) a trainable generative component in form of a convolutional autoencoder (subnetwork I) for adversarial learning; (b) an auxiliary CNN-based gender classifier (subnetwork II); (c) an auxiliary CNN-based face matcher (subnetwork III).

The auxiliary gender classifier as well as the auxiliary matcher\footnote{The term ``auxiliary'' is used to indicate that these subnetworks do not correspond to pre-trained gender classifiers or face matchers, but rather classifiers that are generated from the training data. Note that such a formulation makes the semi-adversarial network generalizable.}  are detachable parts in this network architecture used only during the {\em training} phase. In contrast to GANs, the generative component of this proposed network architecture is a convolutional autoencoder (section~\ref{sec:conv-autoencoder}), which is initially pre-trained to produce an image that closely resembles an image from the training set after incorporating gender prototype information (section~\ref{sec:gender-prototypes}). Then, during further training, feedback from both an auxiliary CNN-based gender classifier and an auxiliary CNN-based face matcher are incorporated into the loss function (see Eqn.~(\ref{eq:loss-function})) to perturb the regenerated images such that the error rate of the auxiliary gender classifier increases while that of the auxiliary face matcher is not unduly affected.

An overview of this semi-adversarial architecture is shown in Fig.~\ref{fig:general-arch}, and the details are further described in the following subsections.

\subsubsection{Convolutional autoencoder} \label{sec:conv-autoencoder}
\begin{figure}[h]
\begin{center}
   \includegraphics[width=1\linewidth]{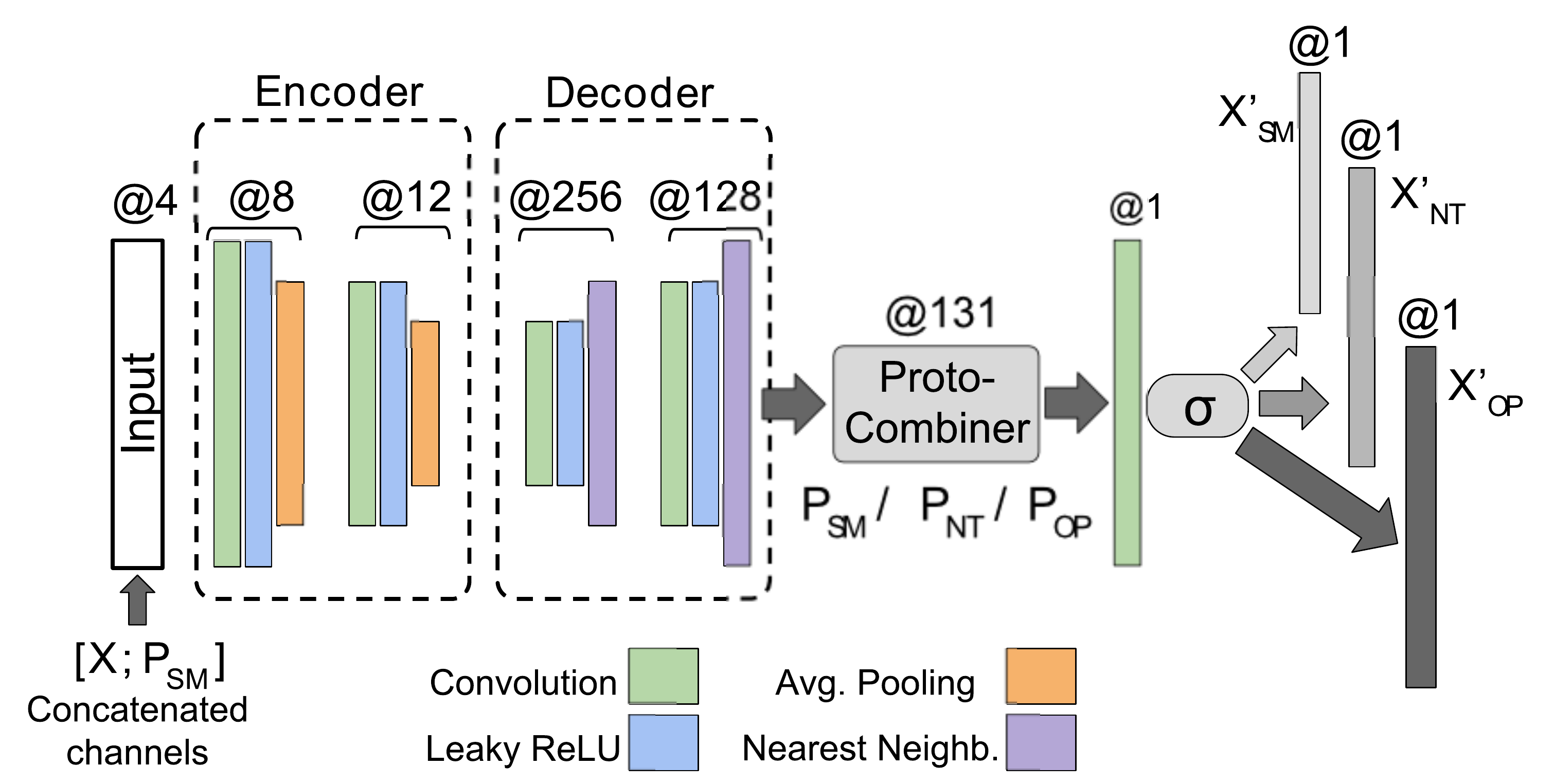}
\end{center}
   \caption{Architecture of the autoencoder augmented with gender-prototype images. The encoder receives a one-channel gray-scale image as input, which is concatenated with the RGB channels of the same-gender prototype image. After the compressed representation is passed through the decoder part of the autoencoder for reconstruction ($128$ channels), the proto-combiner concatenates it with the RGB channels of a same-, neutral-, or opposite-gender prototype resulting in $131$ channels that are then passed to a final convolutional layer. }
\label{fig:conv-autoencoder-arch}
\end{figure}

The architecture of the convolutional autoencoder sub-network that modifies and reconstructs the input image in three different ways is shown in Fig. \ref{fig:conv-autoencoder-arch}. The input to this sub-network is a gray-scale face image of size $224\times 224$  concatenated with a same-gender prototype, $P_{SM}$ (Fig. \ref{fig:gender-prototypes}). The input is then processed through the encoder part consisting of two convolutional layers; each layer is followed by a leaky ReLU activation function and an average pooling layer, resulting in feature maps of size $56\times 56 \times 12$. Next, the outputs of the encoder are passed through a decoder with two convolutional layers each, followed by a leaky ReLU activation and an upsampling layer using two-dimensional nearest neighbor interpolation. The output of the decoder is a $224\times 224\times 128$ dimensional feature map. 

The feature maps from the decoder output are then concatenated with either same-gender ($P_{SM}$), neutral-gender ($P_{NT}$), or opposite-gender ($P_{OP}$) prototypes in the \textit{proto-combiner} module (see Fig.~\ref{fig:conv-autoencoder-arch} and Fig.~\ref{fig:gender-prototypes}). The proto-combiner module is followed by a final convolutional layer and a sigmoid activation function yielding a reconstructed image $X'_{SM}$, $X'_{NT}$, or $X'_{OP}$, depending on the gender-prototype used. The autoencoder described in this section contains five trainable layers. Those layers are pre-trained using an information bottleneck approach~\cite{hinton_reducing_2006} to retain the relevant information from both the original image and the same-gender prototype. This is sufficient to reconstruct realistic looking images by minimizing $J_D(X, X')$, which measures the dissimilarity between the gray-scale input images and the perturbed images by computing the sum of the element-wise cross entropy between input and output (perturbed) images. After pre-training, this subnetwork is further trained by passing its reconstructed images to two  other sub-networks: the auxiliary gender predictor and the auxiliary face matcher (Fig.~\ref{fig:general-arch}). The gender prototypes, as well as the two subnetworks, are described in the following subsections.

\subsubsection{Gender prototypes} \label{sec:gender-prototypes}

\begin{figure}[h]
\begin{center}
   \includegraphics[width=0.68\linewidth]{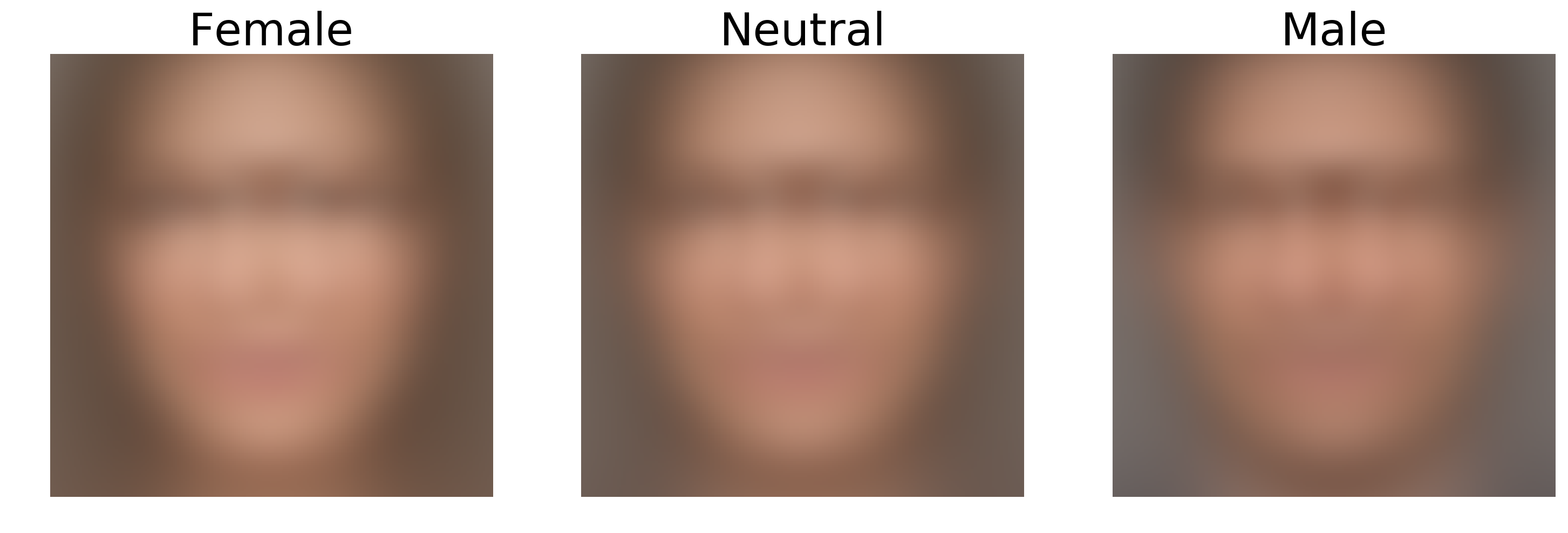}
\end{center}
   \caption{Gender prototypes used to confound gender classifiers while maintaining biometric matching during the semi-adversarial training of the convolutional autoencoder.}
\label{fig:gender-prototypes}
\end{figure}

The $224\times 224$  male and female RGB gender prototypes ($P_\text{male}$, $P_\text{female}$)  were computed as the average of all \num[group-separator={,}]{65160} male images and \num[group-separator={,}]{92190} female images, respectively, in the CelebA training set~\cite{liu_deep_2015}. Then, the same-gender ($P_{SM}$) and opposite-gender ($P_{OP}$) prototypes, which are being concatenated with the input image and combined with the autoencoder output (Fig. \ref{fig:conv-autoencoder-arch}), are constructed based on the ground-truth label $y$, while the neutral-gender prototype is computed as the weighted mean of male and female prototypes (Fig.~\ref{fig:gender-prototypes}):

\begin{itemize}
\item $\text{Same-gender prototype, } P_{SM}\text{:~~} y P_\text{male} + (1-y) P_\text{female}$
\item $\text{Opposite-gender prototype, } P_{OP}\text{:~~} (1-y) P_\text{male} + y P_\text{female}$
\item $\text{Neutral prototype, } P_{NT}\text{:~~} \alpha_F P_\text{female} + \alpha_M P_\text{male}$
\end{itemize}

Here, $\alpha_F$ is the proportion of females in the CelebA training set and $\alpha_M$ is the proportion of males. The convolutional autoencoder network (summarized in Fig.~\ref{fig:general-arch} and further illustrated in Fig.~\ref{fig:conv-autoencoder-arch}) is provided with same-gender prototype images (female or male corresponding to the ground truth label of the input image), which are concatenated with the input image before being transmitted to the encoder module in order to derive a compressed representation of the original image along with the same-gender prototype information. After the decoder reconstructs the original images, the three different gender-prototypes are added as additional channels via the proto-combiner (Fig.~\ref{fig:conv-autoencoder-arch}). 

The final convolutional layer of the  autoencoder produces three different perturbed images: $X'_{SM}$ (obtained  when the same-gender prototype is used), $X'_{NT}$ (when the neutral prototype is used), and $X'_{OP}$ (when the opposite-gender prototype is used). 

\paragraph{Pre-training:} During pre-training, to ensure that the convolutional autoencoder is capable of reconstructing the original images, only the same gender perturbations ($X'_{SM}$) were considered in the cross-entropy cost function. 

\paragraph{Training:} For the further training of the autoencoder, to confound the auxiliary gender classifier and ensure high matching accuracy of the auxiliary matcher, both the perturbed outputs using same- and opposite-gender prototypes were passed through the auxiliary gender classifier, to ensure that the perturbation made using the same-gender prototype produces accurate gender  prediction while perturbations made using the opposite-gender prototype confounds the gender prediction. The perturbed outputs due to the neutral prototypes are not incorporated in the loss function, and are only used for evaluation purposes.

\subsubsection{Auxiliary CNN-based gender classifier} \label{sec:cnn-gender-classifier}

\begin{figure}[h]
\begin{center}
   \includegraphics[width=0.65\linewidth]{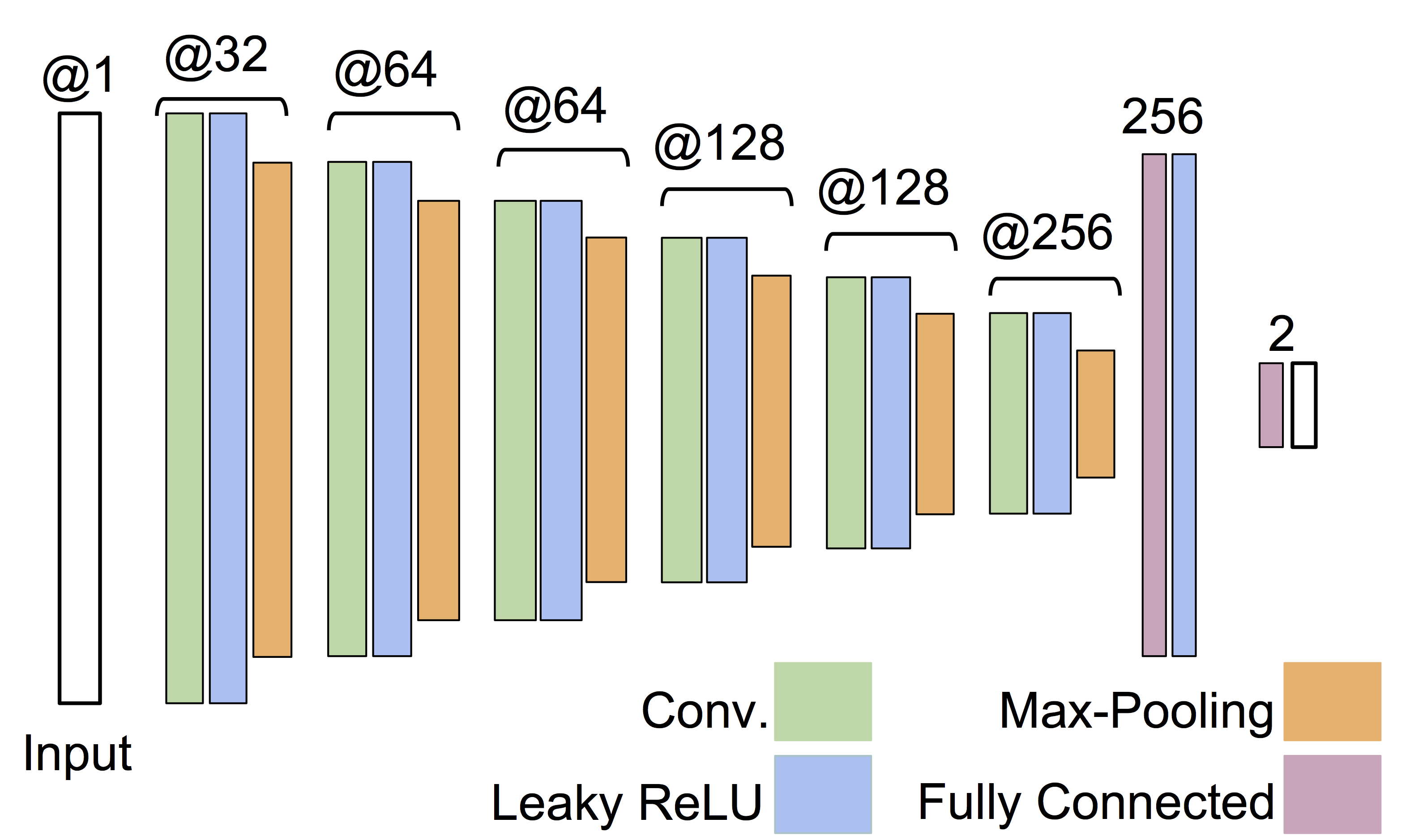}
\end{center}
   \caption{Architecture of the CNN-based auxiliary gender classifier that was used during the training of the convolutional autoencoder. This classifier was  used as an auxiliary (fixed) component in the final model to derive the image perturbations according to the objective function described in Section \ref{sssec:num2}.}
\label{fig:gender-classifier-arch}
\end{figure}

The  architecture of the auxiliary CNN-based gender classifier, which consists of six convolutional layers and two fully connected (FC) layers, is summarized in Fig.~\ref{fig:gender-classifier-arch}. Each convolutional layer is followed by a leaky ReLU activation function and a max-pooling layer that reduces the height and width dimensions by a factor of 2, resulting in feature maps of size $4\times 4\times 256$. Passing the output of the second FC layer through a sigmoid function results in class-membership probabilities for the two labels: $0$:``Female'' and $1$:``Male''. This network was independently trained on the CelebA-train dataset by minimizing the cross-entropy cost function, until its convergence after five epochs; the gender prediction accuracy of the auxiliary network when tested on the CelebA-test set was 96.14\%. During training, two dropout layers with drop probability of $0.5$ were added to the FC layers for regularization. However, these dropout layers were removed when this subnetwork was used for deriving perturbations as part of the three-subnetwork neural network architecture shown in Fig \ref{fig:general-arch}.

As this CNN-based gender classifier was only used for training the convolutional autoencoder for generating perturbed face images, and not for further evaluation of this model, it is referred to as \textit{auxiliary gender classifier} to distinguish it from the gender classifiers used for evaluation.

\subsubsection{Auxiliary CNN-based face matcher} \label{sec:cnn-face-matcher}

As discussed in Section \ref{sssec:num2}, the loss function contains a term $J_M(X, X'; f_M)$ to ensure good face matching accuracy despite the perturbations introduced to confound the gender classifier. To provide match scores during the training of the autoencoder subnetwork, we used a publicly available VGG model as described by Parkhi \etal~\cite{parkhi_deep_2015} consisting of $16$ weight layers. This VGG subnetwork produces face descriptors which are vector representations of size \num[group-separator={,}]{2622} extracted from RGB face images. The publicly available weight parameters of this network were used without further performance tuning. 


In addition, as the open-source VGG-face network expects RGB images as inputs, we modified the convolutional filters of the first layer by adding the three filter matrices related to the input channels, for compatibility with the single-channel gray-scale input images. 
As this CNN-based face matcher was only used for training the convolutional autoencoder for generating perturbed face images, and not for further evaluation of this model, it is referred to as \textit{auxiliary matcher} to distinguish it from the commercial matching software used for evaluation.



\subsection{Loss function}\label{sec:loss-function}

After pre-training the convolutional autoencoder described in Section \ref{sec:conv-autoencoder}, it is connected to the other two subnetworks (the auxiliary CNN-based gender classifier described in Section \ref{sec:cnn-gender-classifier} and the auxiliary CNN-based face matcher described in Section \ref{sec:cnn-face-matcher}) for further training. During the pre-training stage, the loss term $J_D(X, X')$ was used to ensure that the convolutional autoencoder is capable of producing images that are similar to the input images. The loss term is computed as the element- or pixel-wise cross entropy, $S$, between input and output (perturbed) images:

\begin{equation}
J_{D}(X, X'_{SM}) = \sum_{k=1}^{224^2} S\left(X^{(k)} , X'^{(k)}_{SM}\right).
\end{equation}

Next, to generate the perturbed images $X'_{SM}$, $X'_{NT}$, or $X'_{OP}$ (based on the type of gender-prototype used) such that  gender classification is confounded but  biometric matching remains accurate,  two loss terms, $J_G$ and $J_M$, were used. The first loss term is associated with suppressing gender information in $X'_{OP}$ and preserving it in $X'_{SM}$:

\begin{equation}
\begin{array}{cll}
J_G\left(y, X'_{SM}, X'_{OP} ; f_G\right) =& \\ 
\displaystyle S\left(y,  f_G(X'_{SM})\right) +S\left(1-y,  f_G(X'_{OP})\right),
\end{array}
\end{equation}

\noindent where, $S(t, \hat{p})$ denotes the cross-entropy cost function using target label $t$ and the predicted class-membership probability $\hat{p}$. Note that in this loss function, we use the ground truth labels for $X'_{SM}$ so that the gender of $X'_{SM}$ is correctly predicted, while we use flipped labels for $X'_{OP}$  so that the gender of perturbed image $X'_{OP}$ is incorrectly predicted. We found that without the use of this configuration for $X'_{SM}$ and $X'_{OP}$, the network will perturb the input image, $X$, such that perturbations are overfit to the auxiliary gender classifier that is used during training.

The second loss term, $J_M$, measures the matching similarity between input image $X$ and the perturbed image $X'_{SM}$ generated from the same-gender prototype:

\begin{equation}
J_M(X, X'_{SM}; R_{vgg}) = \left\|R_{vgg}(X'_{SM}) - R_{vgg}(X)\right\|_2^2,
\end{equation}
where, $R_{vgg}(X)$ indicates the vector representation of image $X$ obtained from the VGG-face network~\cite{parkhi_deep_2015}.  The total loss is then the weighted sum of the two loss terms $J_G$ and $J_M$:
\begin{equation}
\begin{array}{cll}
J_{total}\left(X, y, X'_{SM}, X'_{OP}; f_G, R_{vgg}\right)=\\ \lambda_{G} J_G(y, X'_{SM}, X'_{OP};f_G) + \lambda_M J_M(X, X'_{SM}; R_{vgg}).
\end{array}
\label{eqn:j_total}
\end{equation}

$J_{total}$ was then used to derive the loss gradients with respect to the parameter weights of the convolutional autoencoder during the training stage, to generate perturbations according to the objective function (Section \ref{sssec:num2}). Note that the coefficients $\lambda_M$ and $\lambda_G$ in Eqn \ref{eqn:j_total} constitute additional tuning parameters to re-weight the contributions of $J_G$ and $J_M$ toward the total loss. In this work, we did not optimize $\lambda_M$ and $\lambda_G$, however, and used a constant of 1 to weight both $J_G$ and $J_M$ equally.

\subsection{Datasets}


The original dataset source used in this work is the large-scale CelebFaces Attributes (CelebA) dataset~\cite{liu_deep_2015}, which consists of \num[group-separator={,}]{202599} face images in JPEG format for which gender attribute labels were already available with  the dataset. The dataset was randomly divided into \num[group-separator={,}]{162079} training images (CelebA-train) and \num[group-separator={,}]{40520} images for testing (CelebA-test). The CelebA-train dataset was used to train the gender classifier (Section \ref{sec:cnn-gender-classifier}), as well as the convolutional autoencoder (Section \ref{sec:conv-autoencoder}).

In addition to the CelebA-test dataset, three publicly available datasets were used for evaluation only: MUCT \cite{milborrow_muct_2010}, LFW \cite{huang_labeled_2007} and  AR-face  \cite{martinezar} databases. The final compositions of these datasets, after applying a preprocessing step using a deformable part model (DPM) as described by Felzenszwalb \etal~\cite{felzenszwalb_cascade_2010} to ensure that all images have the same dimensions ($224 \times 224$), are summarized in Table~\ref{tab:datasets}. The resulting perturbed images obtained from the CelebA-test, MUCT, LFW, and AR-face datasets, were used to measure the effectiveness of modifying the gender attribute as assessed by a commercial gender classifier (G-COTS) and a commercial biometric matcher (M-COTS, excluding AR-images labeled as occluded due to sunglasses or scarfs).

\begin{table}
\caption{Sizes of the datasets used in this study for training and evaluation. CelebA-train was used for training only, while the other four datasets were used to evaluate the final performance of the trained model.}

\label{tab:datasets}
\begin{center}

\scalebox{0.8}{
\begin{tabular}{|l|c|c|c|c|}
\hline
Dataset & Train & \# Images & \# Male & \# Female\\
\hline\hline
CelebA-train & yes & \num[group-separator={,}]{157350}&  \num[group-separator={,}]{65160}& \num[group-separator={,}]{92190}\\
CelebA-test &  no & \num[group-separator={,}]{39411}&  \num[group-separator={,}]{16318}& \num[group-separator={,}]{23093}\\
MUCT &  no &  \num[group-separator={,}]{3754}& 131&145 \\
LFW & no &  \num[group-separator={,}]{12969}& 4205& 1448\\
AR-face & no&  \num[group-separator={,}]{3286}& 76& 60\\
\hline
\end{tabular}}
\end{center}
\end{table}

\subsection{Implementation details and software}

The convolutional autoencoder (Section~\ref{sec:conv-autoencoder}), auxiliary CNN-based gender classifier (Section~\ref{sec:cnn-gender-classifier}) and the auxiliary CNN-based face matcher (Section~\ref{sec:cnn-face-matcher}) were implemented in TensorFlow~\cite{abadi_tensorflow_2016} based on custom code for the convolutional layers and freezing the parameters of the gender classifier and face matcher during training of the autoencoder subnetwork~\cite{raschka_python_2017}.



\section{Experimental Results}

\begin{figure}[t]
\begin{center}
   \includegraphics[width=0.85\linewidth]{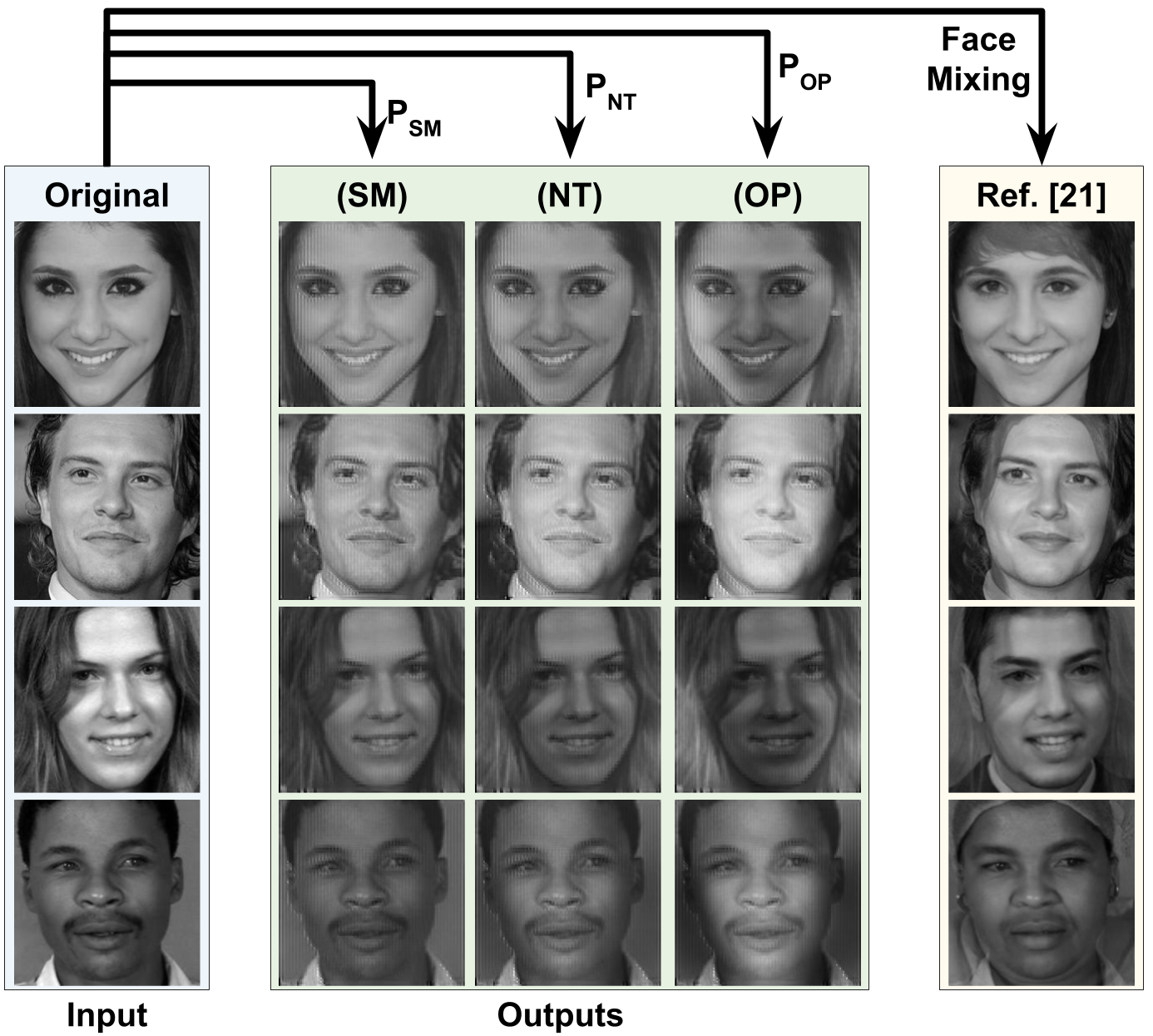}
\end{center}
   \caption{Example input images with their reconstructions using same, neutral, and opposite gender prototypes from the CelebA-test (first two rows) and MUCT (last two rows) datasets.}
\label{fig:example-outputs}
\end{figure}

After training the autoencoder network using the CelebA-train dataset as described in Section \ref{sec:conv-autoencoder}, the model was used to perturb images in other, independent datasets: CelebA-test, MUCT, LFW, and the AR-face database. For each face image in these datasets, a set of three  output images was reconstructed using same-gender, neutral-gender, and opposite-gender prototypes. Furthermore, our results are compared with the face-mixing approach proposed in~\cite{othman_privacy_2014}. Examples of these reconstructed outputs for two female face images, and two male face images are shown in Fig.~\ref{fig:example-outputs}. 

\subsection{Evaluation and verification}

The previously described auxiliary CNN-based gender classifier (Section \ref{sec:cnn-gender-classifier}) and auxiliary CNN-based face matcher (Section \ref{sec:cnn-face-matcher}) were not used for the evaluation of the proposed semi-adversarial autoencoder as these two subnetworks were used to provide semi-adversarial feedback during training. The performance of the semi-adversarial autoencoder is expected to be optimally biased when tested using the auxiliary gender classifier and auxiliary face matcher. 
Thus, we used independent gender classification and face matching software for evaluation and verification instead, to represent a real-world use case scenario. 

Two sets of experiments were conducted to assess the effectiveness of the proposed method. First, two independent software  for gender classification were considered: the popular research software IntraFace~\cite{de_la_torre_intraface_2015} as well as a state-of-the-art commercial software, which we refer to as \textit{G-COTS}. Second, a state-of-the-art commercial matcher that has shown excellent recognition performance on challenging face datasets was used to evaluate the face matching performance; we refer to this commercial face matching software as \textit{M-COTS}.

\subsubsection{Perturbing gender}

\begin{figure}[t!]
\begin{center}
   \includegraphics[width=0.90\linewidth]{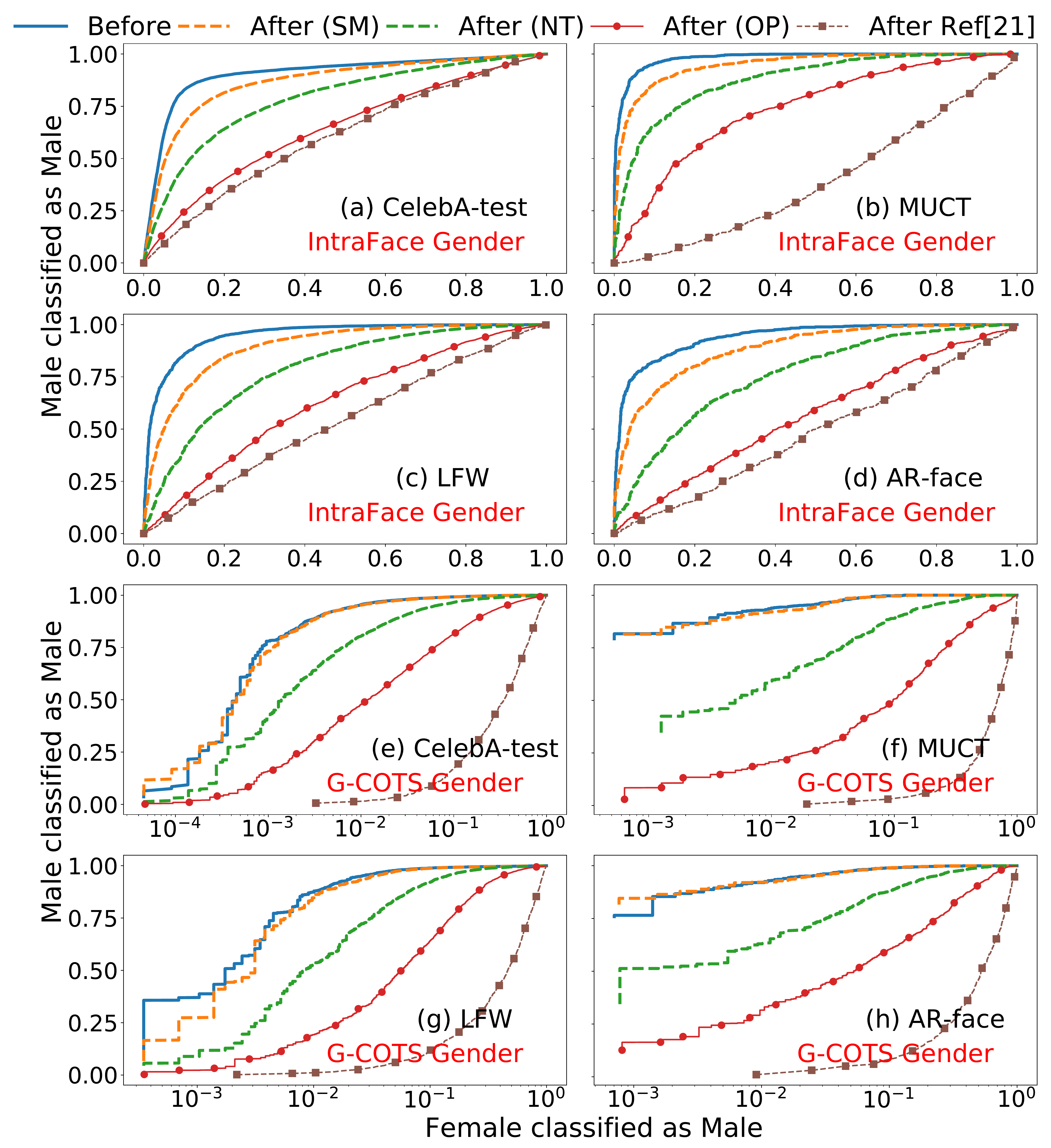}
\end{center}
   \caption{ROC curves comparing the performance of  IntraFace (a-d) and G-COTS (e-h) gender classification software on  original images (``Before'') as well as images perturbed via the convolutional autoencoder model (``After'') on four different datasets: CelebA-test, MUCT,  LFW, and AR-face. }
\label{fig:results-iface}
\end{figure}

In order to assess the effectiveness of the proposed scheme in perturbing gender, the reconstructed images using the proposed semi-adversarial autoencoder from the four datasets were analyzed. The Receiver Operating Characteristic (ROC) curves for predicting  gender  using IntraFace and G-COTS from the original images and the perturbed images are shown in  Fig.~\ref{fig:results-iface}. 

We note that gender prediction via IntraFace is heavily impacted when using different gender prototypes for image reconstruction. We observe that the performance of IntraFace on AR-face images after opposite-gender perturbation is very close to random (as indicated by the near-diagonal ROC curve in Fig.~\ref{fig:results-iface}(a)-(d)). The performance of G-COTS proves to be more robust towards perturbations, compared to IntraFace; however, the ROC curve corresponding to  the opposite-gender prototype, shows a substantial deviation from the ROC curve of the original images (Fig.~\ref{fig:results-iface}(e)-(h)). This observation indicates that the opposite-gender prototype perturbations have a substantial, negative impact on the performance of state-of-the-art G-COTS software, thereby extending gender privacy.  


The exact error rates in predicting the gender attribute of face images using both IntraFace and G-COTS software are provided in Table~\ref{tab:error-rates-gcots} for the original images and the perturbed images using opposite-gender prototypes. The quantitative comparison of the error rates indicates a substantial increase in the prediction error rates when image datasets were perturbed using opposite-gender prototypes. Note that in the case of G-COTS software, perturbations made by the face mixing scheme proposed in~\cite{othman_privacy_2014} result in higher error rates. On the other hand, the additional advantage of our approach is in preserving the identity, as we will see in the next section.

\begin{table}[]
\caption{Error rates in gender prediction using IntraFace and G-COTS gender classification softwares on the original datasets before and after perturbation. Note the substantial increase in the prediction error upon perturbation via the convolutional autoencoder model using opposite-gender prototypes.}
\label{tab:error-rates-gcots}
\begin{center}
\scalebox{0.8}{\begin{tabular}{|l|l|c|c|c|}
\hline
\multirow{2}{*}{Software} & \multirow{2}{*}{Dataset} & Original & Perturbed & \multirow{2}{*}{Ref.~\cite{othman_privacy_2014}} \\
 &  & (before) & (after OP) & \\
\hline\hline
\multirow{4}{*}{IntraFace} & CelebA-test &  19.7\% & 39.3\% & 44.6\%\\
& MUCT &  8.0\% & 39.2\% & 57.7\%\\
& LFW & 33.4\% &  72.5\% & 70.9\%\\
& AR-face & 16.9\%&  53.8\% & 54.2\%\\
\hline
\multirow{4}{*}{G-COTS} & CelebA-test &  2.2\% & 13.6\% &  42.4\%\\
& MUCT &  5.1\% & 25.4\% & 53.9\%\\
& LFW & 2.8\% &  18.8\% & 46.1\%\\
& AR-face & 9.3\%&  26.9\% & 40.6\%\\
\hline
\end{tabular}}
\end{center}
\end{table}

\subsection{Retaining matching accuracy }

\begin{figure}[t!]
\begin{center}
   \includegraphics[width=0.85\linewidth]{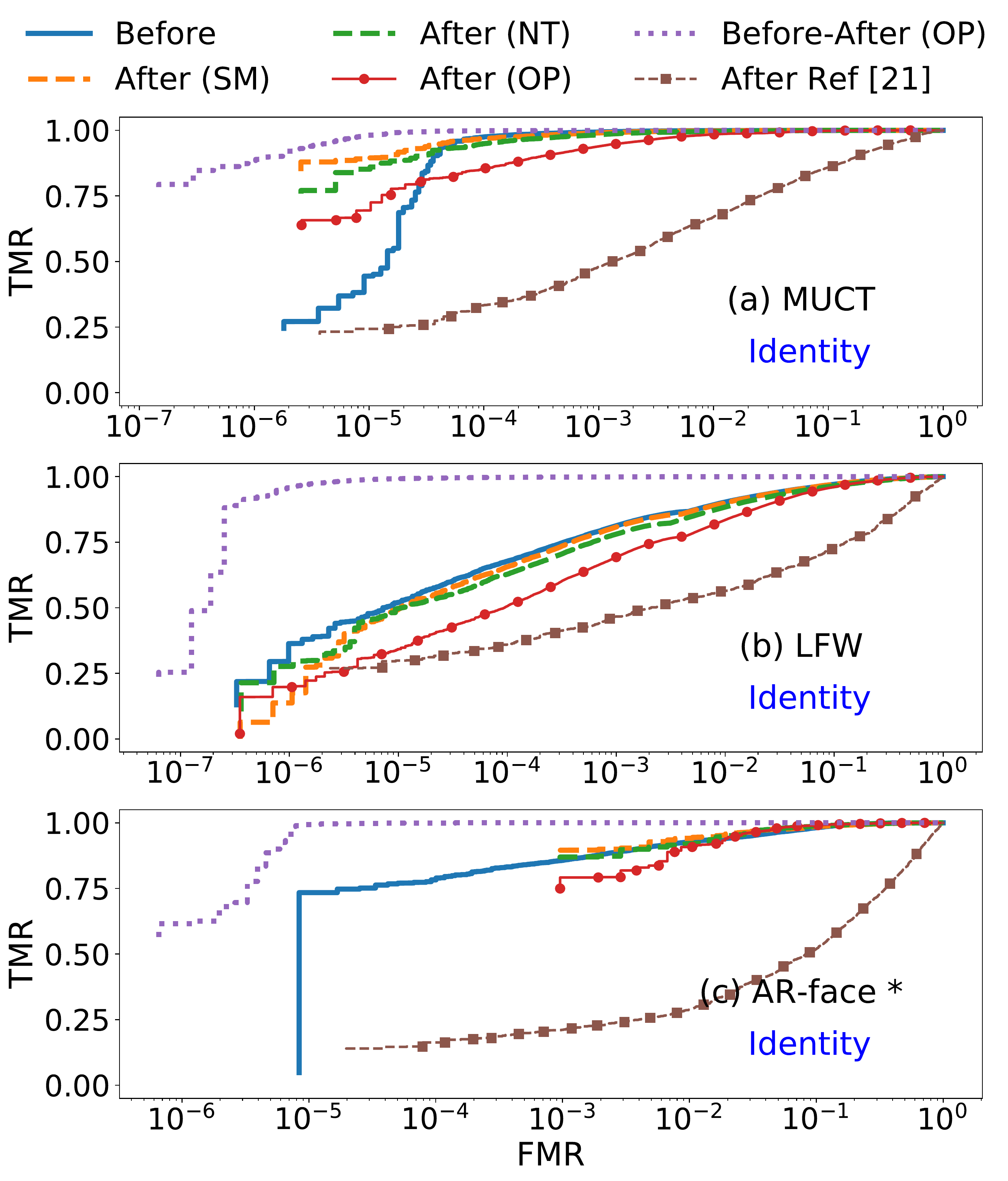}
\end{center}
   \caption{ROC curves showing the performance (true and false matching rates) of M-COTS biometric matching software on the original images (``Before'') compared to the perturbed images (``After'') generated by the convolutional autoencoder model using  same-, neutral-, or opposite-gender prototypes for three different datasets: (a) MUCT, (b) LFW, and (c) AR-face. }
\label{fig:results-mcots}
\end{figure}

The match scores were computed using a state-of-the-art M-COTS software and the resulting ROC curves are shown in Fig.~\ref{fig:results-mcots}. 
While the matching term, $J_M$, in the loss function is directly applied to reconstructed outputs from same-gender prototype, $X'_{SM}$, the reconstructions that use neutral- or opposite-gender prototypes are not directly subject to this loss term (see Section~\ref{sec:loss-function}).  As a result, the ROC curve of the reconstructed images coming from same-gender prototype appear much closer to the original input compared to the reconstructed images from neutral- and opposite-gender prototypes. Overall, we were able to retain a good matching performance even when using opposite-gender prototype. On the other hand, the ROC curves obtained from outputs of the mixing approach proposed in~\cite{othman_privacy_2014} are heavily impacted, resulting in de-identified outputs (which is not desirable in this work).

Finally, the True Match Rate (TMR) values at a False Match Rate of $1\%$ are reported in Table~\ref{tab:tmr-valuess}. The perturbed images from all three datasets show TMR values that are very close to the value obtained from the unperturbed original dataset.

\begin{table}[tbph]
\caption {True (TMR) and false (FMR) matching rates (measured at values of $1\%$) of the independent, commercial M-COTS  matcher after perturbing face images via the convolution autoencoder using same (SM), neutral (NT), and opposite (OP) gender prototypes, indicating that the biometric matching accuracy is not substantially affected by confounding gender predictions.}
\label{tab:tmr-valuess}
\begin{center}
\scalebox{0.8}{
\begin{tabular}{|l||c|ccc|}
\hline
\multirow{2}{*}{Dataset} & Original  & \multicolumn{3}{c|}{Perturbed} \\
&(before) & (SM) & (NT) & (OP) \\
\hline\hline
MUCT &  99.88 \% & 99.79\% & 99.57\% & 98.44\%\\
LFW & 90.29\% & 90.02\% & 88.47\% & 83.45\%\\
AR-face & 94.97\%& 94.11\% & 91.95\% & 90.81\%\\
\hline
\end{tabular}}
\end{center}
\end{table}

\section{Conclusions}

In this work, we focused on developing a semi-adversarial network for imparting soft-biometric privacy to face images. In particular, our semi-adversarial network perturbs an input face image such that gender prediction is confounded while the biometric matching utility is retained. The proposed method uses an auxiliary CNN-based gender classifier and an auxiliary CNN-based face matcher for training the convolutional autoencoder. The trained model is evaluated using two independent gender classifiers and a state-of-the-art commercial face matcher which were unseen during training. Experiments confirm the efficacy of the proposed architecture in imparting gender privacy to face images, while not unduly impacting the face matching accuracy. 

\section{Acknowledgements}

This material is based upon work supported by the National Science Foundation under Grant Number $1618518$.


{\small
\bibliographystyle{ieee}
\balance
\bibliography{egbib}
}

\end{document}